# Automated Generation of Research Workflows from Academic Papers: A Full-text Mining Framework

Heng Zhang[1], Chengzhi Zhang[2,*]

[1]School of Information Management, Central China Normal University, Wuhan 430079, China

[2]Department of Information Management, Nanjing University of Science and Technology, Nanjing 210094, China

**Abstract:** The automated generation of research workflows is essential for improving the reproducibility of research and accelerating the paradigm of "AI for Science". However, existing methods typically extract merely fragmented procedural components and thus fail to capture complete research workflows. To address this gap, we propose an end-to-end framework that generates comprehensive, structured research workflows by mining full-text academic papers. As a case study in the Natural Language Processing (NLP) domain, our paragraph-centric approach first employs Positive-Unlabeled (PU) Learning with SciBERT to identify workflow-descriptive paragraphs, achieving an $F_1$-score of 0.9772. Subsequently, we utilize Flan-T5 with prompt learning to generate workflow phrases from these paragraphs, yielding ROUGE-1, ROUGE-2, and ROUGE-L scores of 0.4543, 0.2877, and 0.4427, respectively. These phrases are then systematically categorized into data preparation, data processing, and data analysis stages using ChatGPT with few-shot learning, achieving a classification precision of 0.958. By mapping categorized phrases to their document locations in the documents, we finally generate readable visual flowcharts of the entire research workflows. This approach facilitates the analysis of workflows derived from an NLP corpus and reveals key methodological shifts over the past two decades, including the increasing emphasis on data analysis and the transition from feature engineering to ablation studies. Our work offers a validated technical framework for automated workflow generation, along with a novel, process-oriented perspective for the empirical investigation of evolving scientific paradigms. Source code and data are available at: https://github.com/ZH-heng/research_workflow.

**Keywords**: Automated research workflow generation; Full-text mining; NLP; Pre-trained language model; Large language model

# 1 Introduction

Rigorous and systematic research workflows form the foundation of reproducible and trustworthy science, constituting a key driver of scientific innovation and advancement (Elliott & Resnik, 2015; Landis et al., 2012). A study is considered reproducible when replicating its research workflow produces results that are consistent with the original findings. The increasing prevalence of paper retractions underscores growing concerns regarding research integrity (Ayodele et al., 2019). Prior studies indicate that irregularities in data processing and analysis are among the primary causes of unreliable results and subsequent paper retractions (Fang et al., 2012; Sharma et al., 2023). These phenomena underscore the critical importance of robust research workflows in ensuring the reproducibility of scientific findings and supporting sustained scientific progress.

A central concept in this study is "*workflow*", which is commonly categorized into "*business workflows*" and "*scientific workflows*". Business workflows refer to the automation of business

---

[*] Corresponding author, email: zhangcz@njust.edu.cn

processes—whether partially or entirely—including the flow of documents, information, and tasks (Choi et al., 2003). Scientific workflows, on the other hand, comprise a series of data processing activities in which data flows sequentially through multiple stages and the dependencies between processes are explicitly defined (Guler et al., 2016; Deelman et al., 2009). The present study primarily focuses on research workflows as presented in academic articles. A research workflow is defined as a systematic sequence of steps and activities undertaken to address a specific scientific problem. For example, a typical research workflow in Natural Language Processing (NLP) is generally structured into three stages: data preparation, data processing, and data analysis. Each stage involves specific tasks such as data annotation, tokenization, model training, and performance evaluation.

Despite its importance, extracting research workflows from academic literature remains a significant challenge. Existing approaches primarily focus on identifying key entities and their relationships, offering only a fragmented representation of the research process. To address this limitation, we propose a novel framework for the automatic generation of comprehensive research workflows through full-text mining of academic papers. Our method visualizes these workflows as flowcharts, aiming not only to provide technical insights but also to transform how researchers engage with scientific knowledge. By mapping the complete structure of research workflows, our framework serves as an innovative tool for literature navigation, offers structured templates to assist novice and interdisciplinary researchers, and establishes a process-oriented foundation for evaluating scientific rigor. These capabilities are particularly vital to the evolving "AI for Science" paradigm (Lu et al., 2024), which calls for standardized, efficient, and unbiased research procedures.

Our approach is informed by Harris's (1952) discourse analysis framework, which regards paragraphs as fundamental units for conveying coherent meaning. Based on this, we design a paragraph-centric mining pipeline. The process begins by identifying and extracting workflow-descriptive paragraphs from academic texts. These paragraphs are then distilled into concise phrases representing individual workflow steps. Finally, the phrases are categorized into three stages—data preparation, data processing, and data analysis—and visualized as unified flowcharts depicting the complete research process.

To operationalize our framework, this study focuses on Natural Language Processing (NLP) as an exemplary domain and investigates three core research questions:

*RQ1*: How can workflow-descriptive paragraphs be reliably identified from full-text papers?

*RQ2*: How can these paragraphs be effectively condensed into workflow phrases?

*RQ3*: How can these workflow phrases be systematically classified into the three stages of research workflows?

To optimize resources utilization, we developed a cost-effective annotation strategy by leveraging structural conventions in NLP papers. Our approach is based on the observation that descriptive paragraph headings frequently indicate workflow-related content, which served as a reliable heuristic for rapidly assembling a large-scale annotated dataset. Using this human-annotated dataset, we concurrently trained models to (1) identify workflow paragraphs (*RQ1*) via Positive-Unlabeled (PU) Learning and (2) generate workflow phrases (*RQ2*). Notably, the models analyze entire paragraph texts rather than solely headings, enabling them to capture richer linguistic patterns characteristic of workflow descriptions. For *RQ3*, we employed the few-shot learning capabilities of ChatGPT[1] to achieve effective classification with minimal annotated examples.

---

[1] https://chatgpt.com

This study makes three primary contributions:

First, we introduce an innovative approach that generates complete research workflows from full-text papers. Unlike prior methods that focus on fragmented procedures, our approach captures the research processes in a holistic format enabling effective visualization of the workflows.

Second, we propose a novel and cost-effective framework that integrates PU Learning, pre-trained models, and large language models such as ChatGPT. This high-performing framework provides a practical and empirically validated method for automated research workflow generation.

Finally, our work introduces a new empirical tool for investigating scientific fields from a process-oriented perspective. By conducting large-scale analyses of research workflows, it uncovers novel insights into the evolution of research paradigms, as demonstrated by our case study on the NLP domain.

The source code and dataset supporting this study are publicly available on GitHub at the following repository: https://github.com/ZH-heng/research_workflow.

## 2 Related Work

The relevant literature encompasses four key areas: research workflow mining, PU Learning classification technique, and the applications of pre-trained models in academic text classification and text generation.

*2.1 Extracting Research Workflows from Academic Papers*

**Table 1. Related works of research workflow extraction from academic papers**

| Authors | Method/Model | Fields | Main Findings |
|---|---|---|---|
| Halioui et al. (2018) | APE rules, WSD | Bioinformatics | Combining ontology and semantic disambiguation could extract workflows from bioinformatics scientific text. |
| Kuniyoshi et al. (2020) | BiLSTM-CRF | Inorganic materials chemistry | BiLSTM-CRF model could effectively identify entities related to battery synthesis and the relationships between them, demonstrating the feasibility of automated synthesis process extraction. |
| Yang et al. (2022) | Manual annotation, NER, RE | Material science | A carefully curated and annotated dataset (PcMSP) with multi-level information is crucial for extracting polycrystalline material synthesis procedures. |
| Ma et al. (2023) | SciBERT, BART, T5 | Natural Language Processing | A structured schema enabled the creation of a knowledge graph by extracting procedural information from scientific publications. |

Workflow mining, also termed process mining or process discovery, is a widely employed technique for analyzing business workflows. It aims to derive meaningful insights from event logs and automatically generate workflow models that illustrate the causal relationships among activities (Aalst, 2009). Within enterprise information systems, business activities are routinely logged and stored as structured data within databases, producing event logs. These logs typically include case IDs, activity names, and start and end timestamps. Each event linked to a specific case and activity within the workflows (Acampora et al., 2017). In contrast, academic papers consist of unstructured text, in which research workflows are primarily described using natural language. This unstructured nature poses considerable challenges for extracting and analyzing research workflows from academic papers.

Despite these challenges, researchers have explored methods including named entity recognition and relation extraction for extracting workflow information from academic papers. Key studies are summarized in **Table 1**. Halioui et al. (2018) introduced an innovative approach to extract workflows from bioinformatics texts, utilizing domain ontologies and word sense disambiguation (WSD) to identify tasks, parameters, data, metadata, and their interconnections. Kuniyoshi et al. (2020) focused on extracting synthesis processes for all-solid-state batteries from scientific literature, employed a BiLSTM-CRF model to detect entities such as materials and operations, and defined three relation types: Condition, Next, and Coreference. Yang et al. (2022) developed PcMSP, a dataset for polycrystalline material synthesis processes, which was derived from 305 open-access papers and includes annotations covering sentences, entities, and relations. Ma et al. (2023) proposed a procedural information extraction framework aimed at optimizing quantitative metrics. They introduced a schema comprising Operation, Effect, and Direction, and applied it to construct a metric-driven mechanism knowledge graph ($MKG_{NLP}$) from large-scale papers.

However, these studies primarily yield fragmented workflow descriptions, failing to capture the full research processes. To address this gap, our study adopts a holistic approach by analyzing full-text academic papers to comprehensively map the complete workflows used to tackle research problems.

*2.2 PU Learning Classification Technique*

Positive-Unlabeled (PU) Learning emerged in the early 2000s to address challenges in semi-supervised document classification. Liu et al. (2002) pioneered this approach by eliminating the requirement for labeled examples in all classes, thus departing from traditional supervised learning paradigms. In their binary classification framework, only the target class ("positive documents") required manual annotation, while the remaining data were treated as unlabeled and predominantly assumed negative. This work established the conceptual foundation for PU Learning. Li and Liu (2003) introduced a two-stage classification strategy for positive and unlabeled samples. Subsequently, Liu et al. (2003) further refined this strategy by conducting a comprehensive evaluation of various methodological combinations and proposed the Biased-SVM algorithm, which demonstrated superior performance. By 2005, PU Learning was formally defined as a binary classification task using positive and unlabeled samples (Li & Liu, 2005).

The two-stage strategy (Liu et al., 2002; Li & Liu, 2003; Kaboutari et al., 2014) remains the most widely adopted approach within PU Learning. It consists of two steps: first, identifying reliable negative samples (RN) from the unlabeled set (U), given a set of labeled positive samples (P), using methods such as the Rocchio algorithm, Bayesian networks, the Spy technique, or 1-DNF; second, training a standard binary classifier with these reliable negatives and labeled positives.

Due to the difficulty in acquiring large, well-labeled datasets, unlabeled samples often vastly outnumber the positive ones in practice. Recent studies have applied PU Learning to various tasks: Chen et al. (2023) used it for weakly supervised noise removal in domain-specific literature datasets, thereby improving data quality without requiring manual annotation. Zhang et al. (2023) developed the SIPUL model for streaming review data, enabling continuous detection of fake reviews. PU Learning has also proven effective for deceptive opinion classification (Hernández Fusilier et al., 2015) and spammer identification (Wu et al., 2020).

*2.3 Academic Text Classification Based on Pre-trained Models*

Text classification, a fundamental task in Natural Language Processing (NLP), involves assigning one or more predefined categories to textual data based on its semantic content. Academic text classification, a subdomain of this task, targets the automated categorization of scholarly documents. Over time, methodologies have progressed from traditional statistical approaches—such as support vector machines and Naive Bayes—to deep learning architectures, and more recently, to pre-trained language models. This evolution reflects an ongoing effort to improve semantic representation, with pre-trained models representing a particularly notable advancement. Trained on large-scale corpora, these models exhibit strong capabilities in capturing linguistic features relevant to diverse downstream tasks. Through task-specific output layers and fine-tuning, researchers can develop classifiers for academic texts that significantly outperform traditional methods. Their ability to capture nuanced semantic relationships makes them especially well-suited for the complexities of scholarly discourse.

General-purpose models such as BERT (Devlin et al., 2019) and RoBERTa (Liu et al., 2019), pre-trained on large open-domain corpora, have achieved outstanding performance across a variety of NLP tasks. However, their performance often declines in domain-specific academic classification tasks due to the lack of specialized domain knowledge in their training data. To address this limitation, domain-adapted models such as SciBERT (Beltagy et al., 2019), FinBERT (Liu et al., 2020), and BioBERT (Lee et al., 2020) have been developed, which are pre-training or fine-tuned on domain-specific corpora. Studies have demonstrated their effectiveness of these models. For example, Ambalavanan and Devarakond (2020) employed SciBERT for multi-criteria classification of biomedical articles, achieving higher precision at comparable recall rates than previous approaches. Likewise, Zhang et al. (2023) utilized SciBERT to classify sentences concerning future work, reporting state-of-the-art accuracy. Additionally, Khadhraoui et al. (2022) developed CovBERT, a model pre-trained on COVID-19 literature, which outperformed RoBERTa, SciBERT, and BioBERT in classifying related scientific texts. Collectively, these advancements underscore the superior efficacy of domain-specific pre-trained models in academic text classification.

*2.4 Text Generation Based on Pre-trained Models*

Text generation, or Natural Language Generation (NLG), is a core task in NLP that seeks to enable machines to produce text that is both grammatically correct and semantically coherent. For example, Roh and Yoon (2023) proposed an LSTM-based model to extract implicit knowledge from scientific literature and generate human-readable statements aimed at identifying potential innovation opportunities. The advent of pre-trained language models (PLMs) has profoundly transformed the field of NLG. Trained on large-scale corpora, these models are capable of capturing complex linguistic patterns and, when fine-tuned, demonstrate strong performance across a wide range of text generation tasks.

Recent studies further exemplify these advancements. Liu et al. (2020) employed a reinforcement learning-based data augmentation strategy with GPT-2 to generate condition-specific text for specialized tasks. Zhang et al. (2020) introduced DIALOGPT, a GPT-2-based dialogue model designed to improve coherence and contextual relevance in conversational settings. Mo et al. (2024) proposed MGCoT, which integrates a Multi-Grained Contexts module into T5 and Flan-T5 to enable table-based text generation, thereby demonstrating improved adaptability to structured data.

Various PLMs optimized for text generation have emerged. For example, BART (Lewis et al.,

2020) employs an Encoder-Decoder architecture and shows versatility across applications. T5 (Raffel et al., 2023) reformulates all NLP tasks in a unified sequence-to-sequence framework, while Flan-T5 (Chung et al., 2022) further refines this approach through via extensive fine-tuning. PEGASUS (Zhang et al., 2020) is designed for abstractive summarization, and MVP (Tang et al., 2023) leverages multi-task pre-training to facilitate knowledge sharing among tasks. Beyond traditional fine-tuning, prompt learning (Lee et al., 2024; Liu et al., 2023) enhances PLMs capabilities by introducing task-specific prompts to improve both understanding and output quality. Collectively, these developments have substantially advanced the capabilities and sophistication of NLG systems.

## 3 Methodology

This study proposes an automated approach to generating research workflows from academic papers. The approach leverages full-text mining techniques in combination with advanced NLP methods, including PU Learning, pre-trained language models, and ChatGPT. As illustrated in **Figure 1**, the proposed framework consists of four key stages. In the first stage, a corpus of NLP research papers in PDF format was compiled. These documents were converted to plain text, parsed to extract relevant information, and manually annotated to create labeled datasets. The second stage involved developing a classification model that combines PU Learning with SciBERT to detect paragraphs containing research workflow descriptions. In the third stage, the Flan-T5 model was fine-tuned on the annotated dataset using prompt learning to produce workflow phrases, which were then automatically generated from relevant paragraphs in unannotated papers. The extracted phrases were subsequently categorized into three research stages—data preparation, data processing, and data analysis—using few-shot prompt learning with ChatGPT. In the final stage, the workflows were organized based on paragraph order and stage classification results, and then visualized as comprehensive flowcharts.

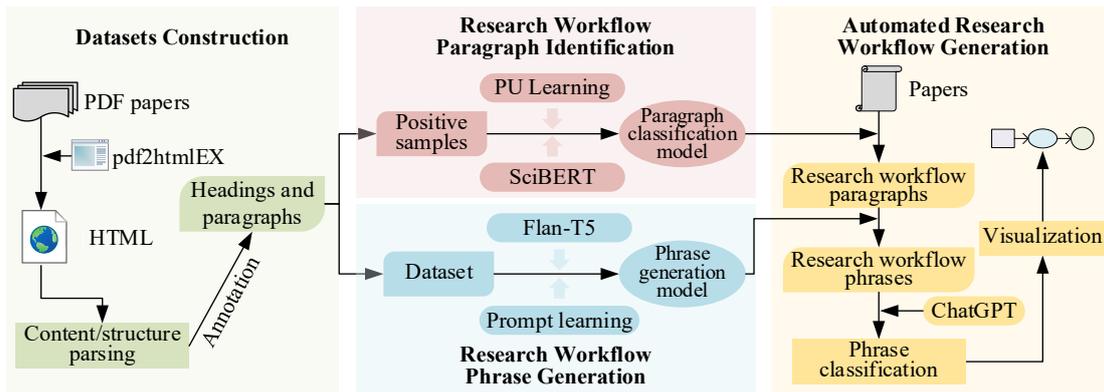

**Figure 1. Framework of this research**

*3.1 Datasets Construction*

This study investigates automated methods for extracting and visualizing research workflows from academic papers, focusing on the NLP domain. Specifically, we analyze papers from three premier conferences: *The Association for Computational Linguistics* (*ACL*), *the Conference on Empirical Methods in Natural Language Processing* (*EMNLP*), and *the North American Chapter of the Association for Computational Linguistics* (*NAACL*). PDF files of these papers were obtained from the ACL Anthology[2]. To ensure data quality, only publications from 2000 to 2022 were included, as earlier PDFs often had lower resolution and lacked detailed workflow descriptions. A total of 17,783 PDFs were converted to HTML using pdf2htmlEX[3], an open-source tool that preserves original document structure and formatting. This conversion enabled the extraction of chapters, headings, and paragraphs based on font characteristics and positional information encoded in HTML tags. **Figure 2** presents the temporal distribution of papers among the three conferences.

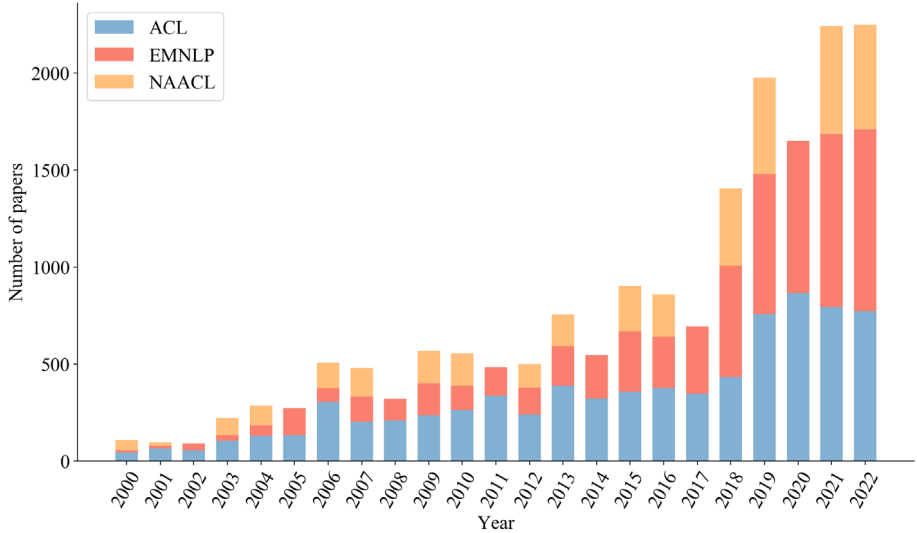

**Figure 2. Annual number of papers published at the ACL, EMNLP, and NAACL conferences (Zhang et al., 2024)**

Our preliminary analysis revealed that NLP researchers often use bold paragraph headings to highlight key methodological steps (see **Figure 3**). To efficiently construct a large-scale annotated dataset, we developed a heuristic leveraging this common scholarly writing convention. We hypothesized that paragraphs with descriptive headings (e.g., "Data Collection," "Model Training") are most likely to contain detailed workflow descriptions. This strategy does not require all workflow steps to be explicitly stated in headings. Instead, it uses the presence of such titles as a practical and scalable method to identify candidate paragraphs for annotation.

Using a rule-based extraction method, we collected 25,179 titled paragraphs from 5,000 academic papers. Each paragraph underwent a two-stage annotation process: first, to determine whether the heading independently denoted a research workflow step; and second, to verify whether the paragraph content accurately elaborated the step implied by the heading. Paragraphs meeting both criteria were labeled as positive ("1"), while all others were labeled as negative ("0"). This process yielded 7,761 positive instances out of 25,179. To evaluate annotation quality, 20% of the corpus was randomly sampled for inter-annotator agreement (IAA) analysis. Two groups of trained annotators independently labeled this subset, achieving a Cohen's Kappa coefficient of 0.70, indicating substantial agreement. Detailed annotation guidelines are available at: https://github.com/ZH-heng/research_workflow/blob/main/Annotation_Guidelines.docx.

This approach enabled the efficient construction of two high-quality datasets from a single annotation effort. Specifically, **For Paragraph Identification (*RQ1*),** the 7,761 positively-labeled paragraphs serve as the positive set (P) for PU Learning. **For Phrase Generation (*RQ2*),** the corresponding (*paragraph_text, workflow_phrase*) pairs from these 7,761 instances form the dataset for training and evaluating our generation models. Crucially, the models are trained on full paragraph contents, allowing them to capture both semantic and discursive characteristics of workflow descriptions. As a result, the models are able to identify and summarize workflow steps during inference, even when processing paragraphs lacking explicit descriptive headings.

---

[2] https://aclanthology.org
[3] https://pdf2htmlex.github.io/pdf2htmlEX

**Figure 3. Exemplar paragraph heading functioning as research workflow phrase with corresponding methodological description in NLP paper**

*3.2 Research Workflow Paragraph Identification Based on PU Learning and Pre-trained Models*

This study employed PU Learning combined with pre-trained models to identify research workflow paragraphs in academic papers, utilizing the 7,761 annotated workflow paragraphs as positive samples. A primary challenge in PU Learning is the identification of reliable negative samples. We selected the "Spy technique" for its simplicity, effectiveness, and flexibility. As shown in **Figure 4**, this technique involves inserting a small random subset of known positive samples ("Spy" samples) into the unlabeled dataset. A classifier is then trained to distinguish the remaining positive samples from this combined set, and the predicted probabilities of the "Spy" samples are used to set a threshold for identifying reliable negative samples within the unlabeled data.

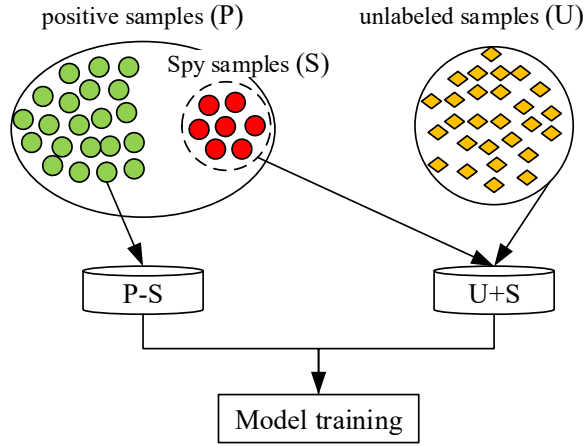

**Figure 4. Data processing pipeline of "Spy technique"**

In our implementation, 15% of the positive set (P) was randomly selected to form the Spy set (S), which was then inserted into a large unlabeled set (U) comprising 77,610 paragraphs randomly sampled from unannotated papers. A fastText[4] classifier was trained using the remaining positive instances (P–S) as the positive class and the combined set (U + S) as the negative class. Following Li and Liu (2003), and to mitigate potential noise in the unlabeled data, we set the decision threshold at the 1st percentile of predicted probabilities for U. This yielded 9,445 candidate negative samples, from which the 7,761 instances with the lowest probability scores were selected as the reliable negative (RN) set, ensuring class balance for subsequent training.

The final dataset for classification, comprising the positive (P) and reliable negative (RN) sets, was partitioned into training, validation, and test sets with an 8:1:1 ratio. We then developed PLM-based binary classifiers to identify research workflow paragraphs. As depicted in **Figure 5**, the model comprises these core components:

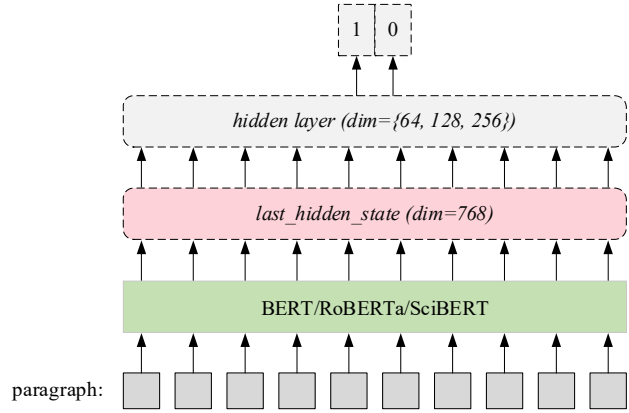

**Figure 5. Research workflow paragraph classifier based on pre-trained language model**

**Input Layer:** Paragraphs ($P$) are tokenized into sequences ($X = [x_1, \ldots, x_n]$) for PLM processing.

**PLM Layer:** Utilizing BERT, RoBERTa, or SciBERT (PLM), this layer extracts contextualized representations. The last hidden state ($H_{last\_hidden\_state}$) of the PLM, with a dimension of 768, serves as the paragraph embedding:

$$H_{last\_hidden\_state} = PLM(X) \qquad (1)$$

**Hidden Layer:** A fully connected hidden layer with *ReLU* activation further refines the representation:

$$H_{hidden\_layer} = ReLU(W_{hidden} H_{last\_hidden\_state} + b_{hidden}) \qquad (2)$$

We evaluated hidden layer sizes of 64, 128, and 256 neurons to optimize performance.

**Output Layer:** A binary classification layer outputs a probability ($y_{prob}$) via a sigmoid function ($\sigma$):

$$y_{prob} = \sigma(W_{output} H_{hidden\_layer} + b_{output}) \qquad (3)$$

The final classification ($y_{class}$) is determined by a 0.5 threshold:

$$y_{class} = \begin{cases} 1, & \text{if } y_{prob} \geq 0.5 \\ 0, & \text{if } y_{prob} < 0.5 \end{cases} \qquad (4)$$

The proposed model employs pre-trained contextual embeddings, followed by a task-specific hidden layer and a binary output layer, to achieve efficient classification of research workflow paragraphs. We systematically evaluated different hidden layer dimensionalities to determine the optimal configuration. For comparison, we also implemented three traditional baseline models: SVM (Support Vector Machine), Random Forest, and XGBoost (eXtreme Gradient Boosting).

---

[4] https://fasttext.cc

*3.3 Research Workflow Phrase Generation Based on Pre-trained Models and Prompt Learning*

To generate research workflow phrases, we constructed a dataset comprising 7,761 pairs of paragraph text and their corresponding annotated headings, which serve as workflow phrases. This dataset was divided into training, validation, and test sets in an 8:1:1 ratio.

Our methodology employed a two-stage optimization process. First, we fine-tuned five pre-trained text generation models (BART, T5, Flan-T5, PEGASUS, and MVP) to determine the most effective model architecture, evaluating performance using ROUGE-1, ROUGE-2, and ROUGE-L metrics (Lin, 2004). Second, we applied prompt learning to the best-performing model to further enhance phrase generation quality. Prompt learning leverages task-specific input prompts to guide the model toward more accurate outputs (He et al., 2024; Zhu et al., 2024). Recognizing that Flan-T5 is a powerful instruction-tuned model optimized to follow direct commands effectively, our goal was not to develop a novel prompting technique, but rather to systematically identify the most effective instruction to unlock the model's potential for our workflow phrase generation task. Accordingly, instead of designing overly complex prompt structures, we focused on exploring a design space of concise templates to empirically determine the most effective prompt. We designed ten prompt templates (see **Table 2**) containing placeholders for paragraph content ("[X]") and target phrases ("[Y]"). These prompts were grouped three groups: (1) Prompts 1–2 without initial task verbs; (2) Prompts 3–7 starting with "Summarize"; and (3) Prompts 8–10 beginning with "Generate". All prompts combine "title" or "research workflow" designations with core instructions to guide the model's focus.

Table 2. Manually designed prompt templates

| ID | Prompt template |
| --- | --- |
| Prompt 1 | [X] The title of the above text is: [Y] |
| Prompt 2 | [X] The title describing research workflow of the above text is: [Y] |
| Prompt 3 | [X] Summarize a title for the above text: [Y] |
| Prompt 4 | [X] Summarize a title describing research workflow for the above text: [Y] |
| Prompt 5 | [X] Summarize the research workflow described in the above text: [Y] |
| Prompt 6 | [X] Summarize a research workflow from the above text: [Y] |
| Prompt 7 | [X] Summarize the above text with a phrase: [Y] |
| Prompt 8 | [X] Generate a title for the above text: [Y] |
| Prompt 9 | [X] Generate a title describing research workflow for the above text: [Y] |
| Prompt 10 | [X] Generate a research workflow from the above text: [Y] |

*3.4 Research Workflow Phrase Classification Using ChatGPT with Few-Shot Learning*

Based on a detailed analysis of NLP research and extensive annotation of research workflows, we proposed a three-stage framework (data preparation, processing, and analysis) to systematically organize NLP research workflows. The generated workflow phrases are then categorized according to these stages, defined as follows:

**Data Preparation:** This stage involves the acquisition, refinement, and organization of data,

including sourcing raw datasets, performing preprocessing tasks such as cleaning and formatting, constructing annotated corpora, providing preliminary statistical descriptions, and splitting data into training, validation, and test sets.

**Data Processing:** During this stage, various operations are applied to the prepared dataset, such as information extraction, model development and training, as well as data augmentation, to facilitate subsequent analysis.

**Data Analysis:** This stage focuses on interpreting processed data to derive meaningful insights. It encompasses model evaluation (manual or automated), comparative analysis, case studies, correlation analysis, and assessment of factor impacts.

Given the broad scope of data processing and analysis, future research may further refine these stages and reclassify workflow phrases as appropriate. When adapting this framework to other domains, domain-specific characteristics should inform its design.

Classifying workflow phrases into three stages can be treated as a supervised learning task, which conventionally requires a labeled dataset. However, manual annotation is often prohibitively expensive and time-consuming. To reduce annotation costs, we employed ChatGPT for automated classification, leveraging few-shot prompting strategy with category-specific exemplars to enhance accuracy.

Our methodology involved designing a detailed prompt to guide the large language model. The prompt was optimized for both clarity and efficiency through three key elements: (1) contextual framing established ChatGPT's role as an "NLP expert" to ensure domain-appropriate responses; (2) providing five carefully curated examples per category to help ChatGPT disambiguate the three research stages (data preparation, data processing, and data analysis); and (3) clear instructions for output formatting (*original phrase // label*) to ensure structured, machine-readable responses.

This process was implemented via the ChatGPT API using the "gpt-4o-mini" model. To ensure deterministic and reproducible outputs, the *temperature* parameter was set to 0, while other parameters retained their default values. The classification was executed in batches, with each API call processing ten research workflow phrases.

## 4 Experimental Results and Analysis

This chapter evaluates methods for identifying, generating, and classifying research workflow components. We describe the experimental setup, present and analyze results for each task, and explore the automatically generated research workflows from NLP papers.

*4.1 Experimental Setup*

We employed pre-trained language models from Hugging Face[5]. Common settings across all experiments included training for five epochs with validation after each epoch to select the best-performing checkpoint for final evaluation. For research workflow paragraph classification, we used the "base" variants of BERT, RoBERTa, and SciBERT with a batch size of 64 and learning rates from 1e-5 to 5e-5. For workflow phrase generation, we adopted the "large" variants of BART, MVP, PEGASUS, T5, and Flan-T5. Due to memory limitations, we adjusted the batch sizes to 16 for BART, MVP, and PEGASUS, while further reducing them to 8 for T5 and 6 for Flan-T5. The learning rates for these text generation models were systematically evaluated within the range of 3e-5 to 7e-5. The maximum sequence length (max_len) was set at the 90th percentile of token lengths. Experiments were conducted on a Windows 10 system with an Intel Core i9-13900KF CPU and an NVIDIA RTX 4090 GPU (24GB).

---
[5] https://huggingface.co

*4.2 Evaluation Metrics*

We assessed our approach using established evaluation metrics for each task, as detailed below.

4.2.1 Evaluation Metrics for Research Workflow Paragraph Classification

In PU Learning scenarios, the absence of labeled negative samples necessitates a focus on the prediction performance for positive samples, which typically represent the class of interest (Li & Liu, 2005). Accordingly, we evaluated classification performance using Precision (P), Recall (R), and $F_1$-score for predicted positive samples. The formulas are as follows:

$$P = \frac{TP}{TP + FP} \tag{5}$$

$$R = \frac{TP}{TP + FN} \tag{6}$$

$$F_1 = \frac{2 \times P \times R}{P + R} \tag{7}$$

Where:
- **TP (True Positive):** The count of research workflow paragraphs correctly identified as belonging to the target category.
- **FN (False Negative):** The count of research workflow paragraphs erroneously classified as non-research workflow paragraphs.
- **FP (False Positive):** The count of non-research workflow paragraphs mistakenly identified as research workflow paragraphs.
- **TN (True Negative):** The count of non-research workflow paragraphs accurately recognized as non-target instances.

4.2.2 Evaluation Metrics for Research Workflow Phrase Generation

To assess the quality of the generated research workflow phrases, we employed three standard automatic evaluation metrics: ROUGE-1, ROUGE-2, and ROUGE-L. ROUGE-1 and ROUGE-2 measure the overlap of unigrams and bigrams between generated and reference texts, while ROUGE-L evaluates the longest common subsequence (LCS) between the texts, capturing the longest sequence of words that appears in both texts while maintaining order. These metrics are calculated as follows:

$$ROUGE - N = \frac{\sum_{S \in \{ReferenceHeadings\}} \sum_{N\_gram \in S} Count_{match}(N\_gram)}{\sum_{S \in \{ReferenceHeadings\}} \sum_{N\_gram \in S} Count(N\_gram)} \tag{8}$$

For this study, the generated text is defined as the research workflow phrases produced by our models, and the reference text as the paragraph headings in the test set. ROUGE-1 and ROUGE-2 are calculated using Equation (4) when $N$ equals 1 and 2, respectively. In this context, $Count_{match}(N\_gram)$ is the number of $N$-grams co-occurring in both generated and reference texts, and $Count(N\_gram)$ is the total number of $N$-grams in the reference text.

ROUGE-L evaluates the similarity between generated text and reference text based on the longest common subsequence (LCS). The subsequent formula details how ROUGE-L is computed:

$$P_{ROUGE-L} = \frac{\sum_{i=1}^{n} LCS(GeneratedPhrase_i, ReferenceHeading_i)}{\sum_{i=1}^{n} Count(word, GeneratedPhrase_i)} \qquad (9)$$

$$R_{ROUGE-L} = \frac{\sum_{i=1}^{n} LCS(GeneratedPhrase_i, ReferenceHeading_i)}{\sum_{i=1}^{n} Count(word, ReferenceHeading_i)} \qquad (10)$$

$$F_{ROUGE-L} = \frac{(1+\beta^2) * P_{ROUGE-L} * R_{ROUGE-L}}{P_{ROUGE-L} + \beta^2 R_{ROUGE-L}} \qquad (11)$$

Where, $LCS(GenerationPhrase_i, ReferenceHeading_i)$ represents the length of the longest common subsequence between the *i*-th generated workflow phrase and its corresponding original paragraph heading. $Count(word, GeneratedPhrase_i)$ and $Count(word, ReferenceHeading_i)$ denote the frequency of the specified word within the *i*-th generated workflow phrase and the corresponding reference heading, respectively. Finally, the F$_1$ with $\beta$ set to 1, is used to quantify the ROUGE-L score.

*4.3 Performance Evaluation of Research Workflow Component Identification, Generation and Classification*

This section systematically assesses the effectiveness of our proposed framework in three key aspects: (1) identifying paragraphs containing research workflow descriptions, (2) generating research workflow phrases, and (3) classifying identified workflow phrases using ChatGPT.

4.3.1 Identification of Research Workflow Paragraph

Following the two-stage PU learning methodology detailed in Section 3.2, we trained and evaluated three PLM-based classifiers and three baseline models to address *RQ1*. The experimental results are presented in **Table 3**.

Table 3. Evaluation results of research workflow paragraph classification

| Model | P | R | F$_1$ |
|---|---|---|---|
| SVM | 0.9161 | 0.909 | 0.9125 |
| Random Forest | 0.8613 | 0.8882 | 0.8745 |
| XGBoost | 0.9099 | 0.9064 | 0.9081 |
| BERT | 0.9699 | 0.9636 | 0.9667 |
| RoBERTa | 0.9551 | 0.9675 | 0.9612 |
| SciBERT | **0.9791** | **0.9753** | **0.9772** |

SciBERT consistently outperformed all other evaluated models in identifying research workflow paragraphs, achieving a precision of 0.9791, a recall of 0.9753, and an F$_1$-score of 0.9772. This performance represents a significant improvement over the baseline models. Specifically, compared to BERT, SciBERT showed increases of 0.92% in precision, 1.17% in recall, and 1.05% in F$_1$-score. The gains over RoBERTa were 2.40%, 0.78%, and 1.60% for the same metrics, respectively. These results underscore SciBERT's superior efficacy for this classification task. SciBERT's advantage can be primarily attributed to its pre-training on a specialized corpus comprising 1.14 million scientific papers from the biomedical and computer science domains. In contrast to the general-purpose BERT and RoBERTa, SciBERT leverages prior knowledge tailored to the scientific lexicon and discourse structure. This specialized foundation enhances its ability to make fine-grained distinctions between positive and negative samples, thereby improving classification accuracy.

### 4.3.2 Generation of Research Workflow Phrases

To address *RQ2*, we evaluated the models for workflow phrase generation as described in Section 3.3. The evaluation proceeded in two stages: first, selecting the best-performing base model, and second, optimizing it with prompt learning. The complete experimental results are presented in **Table 4**.

Table 4. Evaluation of research workflow phrase generation

| Stage | Model | ROUGE-1 | ROUGE-2 | ROUGE-L |
|---|---|---|---|---|
| **Base Model Selection** | BART | 0.4232 | 0.2647 | 0.4115 |
| | MVP | 0.4266 | 0.2623 | 0.4161 |
| | PEGASUS | 0.4020 | 0.2533 | 0.3936 |
| | T5 | 0.4347 | 0.2657 | 0.4222 |
| | Flan-T5 | **0.4434** | **0.2754** | **0.4322** |
| **Prompt-based Optimization** | Flan-T5+Prompt 1 | 0.4515 | 0.2812 | 0.4388 |
| | Flan-T5+Prompt 2 | 0.4480 | 0.2790 | 0.4359 |
| | Flan-T5+Prompt 3 | 0.4488 | 0.2771 | 0.4364 |
| | Flan-T5+Prompt 4 | 0.4512 | 0.2862 | 0.4407 |
| | Flan-T5+Prompt 5 | 0.4434 | 0.2758 | 0.4323 |
| | Flan-T5+Prompt 6 | 0.4469 | 0.2832 | 0.4367 |
| | Flan-T5+Prompt 7 | 0.4465 | 0.2836 | 0.4382 |
| | Flan-T5+Prompt 8 | 0.4456 | 0.2811 | 0.4342 |
| | Flan-T5+Prompt 9 | **0.4543** | **0.2877** | **0.4427** |
| | Flan-T5+Prompt 10 | 0.4350 | 0.2760 | 0.4240 |

In the base model selection stage, Flan-T5 exhibited superior performance, achieving the highest scores across all evaluation metrics: 0.4434 (ROUGE-1), 0.2754 (ROUGE-2), and 0.4322 (ROUGE-L). Notably, both Flan-T5 and its close competitor T5, which share similar Transformer-based encoder-decoder architectures, significantly outperformed the other models. This performance advantage stems from their unified text-to-text framework. While T5 benefits from extensive pre-training, Flan-T5 incorporates critical improvements, including optimized data curation, Chain-of-Thought fine-tuning, and expanded model scale, which collectively enhance its semantic understanding and generation capabilities.

Building on these findings, we conducted prompt-based optimization to maximize Flan-T5's performance. As detailed in the second section of **Table 4**, this approach significantly improvements in phrase generation. Notably, the Flan-T5 model with Prompt 9 achieved optimal performance across all three metrics: 0.4543 (ROUGE-1), 0.2877 (ROUGE-2), and 0.4427 (ROUGE-L), representing a substantial improvement over the baseline Flan-T5 model. Furthermore, Flan-T5 paired with Prompt 1 and Prompt 4 also exhibited strong performance, both achieving ROUGE-1 scores above 0.45. To validate the robustness of our approach, we selected the top-performing Flan-T5 + Prompt 9 model for further analysis. We conducted five experiments using different random seeds, and a paired-samples t-test confirmed that its performance gains over the baseline (Flan-T5 mode) are statistically significant across all metrics (ROUGE-1, ROUGE-2, and ROUGE-L), with all p-values < 0.05.

A comparative analysis of the prompts reveals the nuances of prompt design. Prompt 9 ("Generate a title describing research workflow...") outperformed Prompt 4 ("Summarize a title describing research workflow..."), suggesting that the verb "Generate" may be more effective than "Summarize" for this task. However, this was not universally true; Prompt 3 ("Summarize a title...") outperformed Prompt 8 ("Generate a title..."), highlighting that the subsequent text in the prompt is critical in guiding the model. Conversely, Flan-T5 with Prompt 10 ("Generate a research workflow...") underperformed the baseline, while Prompt 5 ("Summarize the research workflow...") only matched it. Both prompts omitted the word "title," which appears to be essential. Since the task was to generate title-like phrases, the term "title"—a common word in Flan-T5's training data—likely provided crucial context that the less-frequent term "research workflow" alone could not. Overall, these results validate our pragmatic approach. The statistically significant performance improvement achieved using a simple, direct instruction (Prompt 9) confirms that, for a well-defined task and a capable instruction-tuned model, systematically identifying the optimal prompt phrasing serves as a highly effective optimization strategy.

4.3.3 Classification of Research Workflow Phrases

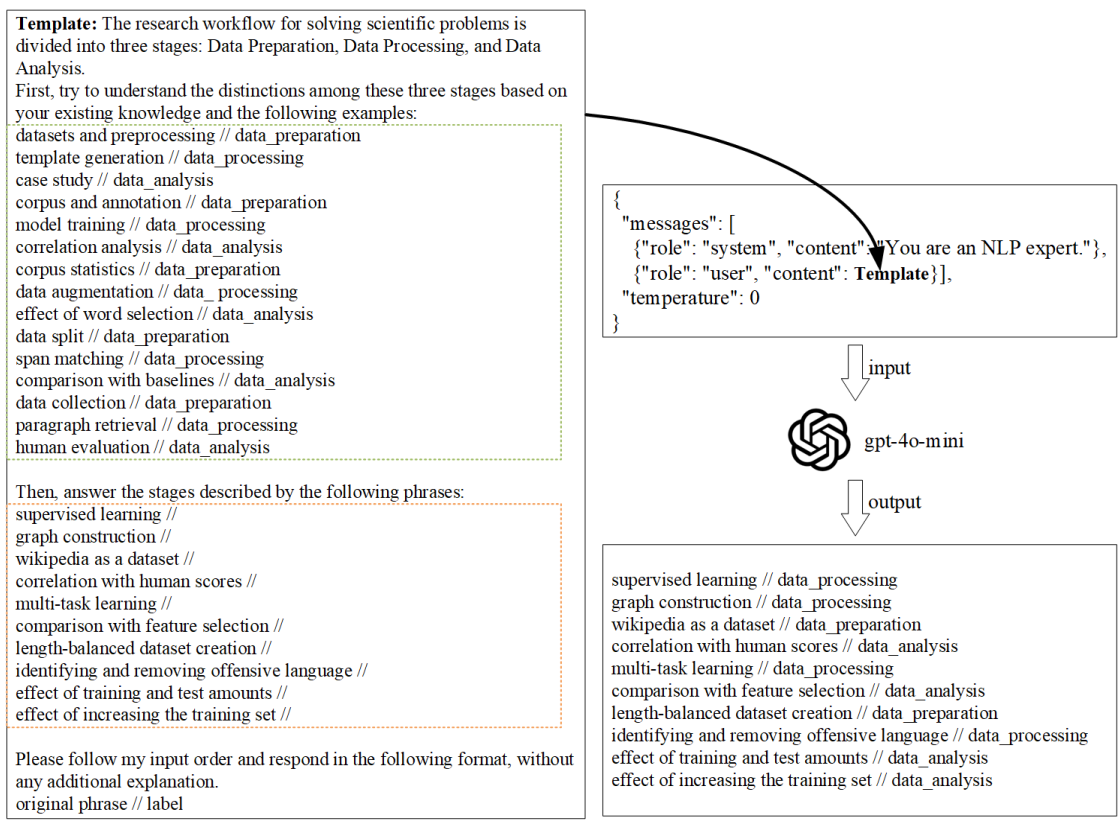

**Figure 6. Workflow phrase classification via ChatGPT with few-shot learning**

This study developed a framework for extracting research workflows from academic publications. Using NLP papers as a case study, we processed 17,783 PDF documents. A SciBERT-based classifier first identified workflow-descriptive paragraphs, followed by phrase generation using the Flan-T5+Prompt 9 model. This yielded 227,814 unique phrases from 17,725 papers (58 papers contained no identifiable workflow content). After lemmatization and clustering based on edit distance (>0.9), we retained the most frequent variant from each cluster, resulting in 197,304 standardized workflow phrases. To address RQ3, these phrases were categorized into three stages—data preparation, data processing, and data analysis—via few-shot prompting with ChatGPT, as detailed in Section 3.4.

**Figure 6** illustrates the complete classification process for a sample batch of phrases. It showcases the detailed prompt—including the system role, few-shot exemplars, and input phrases—and the corresponding structured output from the "gpt-4o-mini" model. As shown in the figure, the model correctly assigns labels like "*data_preparation*," "*data_processing*" and "*data_analysis*," demonstrating its strong capability to discern the nuances of each workflow stage based on the provided examples. To further validate these findings, we manually examined a random subset of 1,000 classified instances, yielding a precision score of 0.958. The result confirms the effectiveness of few-shot learning with ChatGPT for workflow phrase categorization.

*4.4 Automated Generation and Visualization of Structured Research Workflows for NLP Papers*

We have derived structured research workflows for NLP papers by identifying workflow-descriptive paragraphs, generating concise workflow phrases based on paragraph contents, and categorizing these phrases into three stages: data preparation, data processing, and data analysis. **Table 5** presents workflow samples extracted from several papers.

Table 5. Structured research workflows derived from several papers

| Paper_id | Data Preparation | Data Processing | Data Analysis |
|---|---|---|---|
| 2021.emnlp-main.443 | datasets and preprocessing | document retrieval, multiple [cls] embeddings, projection-based pooling, pre-training, training setup, query pooling, joint pre-training | cost of model deployment, effect of length of document representations, cost of projections, effect of projection heads, improved re-ranking with mva |
| 2021.emnlp-main.33 | datasets and preprocessing | text clustering, som processing, converting feature layer to mono-dimensional index, lstm representation, chinese bert-based model | effectiveness of our proposed somncscm, data and evaluation, automatic evaluation, effect of cluster index features, ablation study, clustering performance, case study, error analysis |
| 2020.acl-main.258 | data collection, gold standard creation, baselines and preprocessing | contrastive term extraction, feature selection, centering and batch normalization, comparative embeddings and multi-channel model, network training | classifying technicality, qualitative analysis, ambiguous terminology detection, degree of technicality detection, effect of centering, evaluation and comparison, vector comparison |

We then used the Mermaid[6] diagramming tool to automatically generate visual representations of research workflows. As illustrated in **Figure 7** (based on paper "2021.emnlp-main.443"), the typical NLP research workflow is structured into three sequential stages: data preparation → data processing → data analysis. Within each stage, sub-workflows are arranged in the order of their corresponding paragraphs in the source paper, with directional arrows indicating inter-step dependencies.

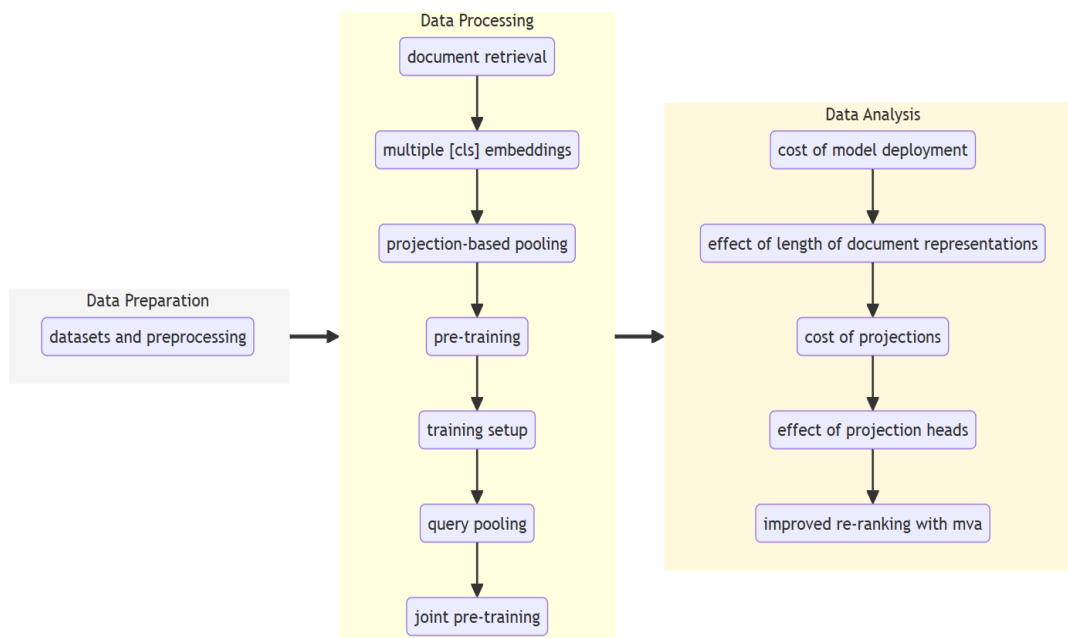

**Figure 7. Visualization of automatically generated research workflow**

**Figure 7** presents the complete workflow of the paper *"Multi-Vector Attention Models for Deep Re-ranking"*. The study follows a structured pipeline starting with "datasets and preprocessing" under the data preparation stage. The data processing stage includes seven ordered steps, beginning with "document retrieval" and "multiple [CLS] embeddings", indicating the generation of diverse document representations. This is followed by "projection-based pooling", application of "pre-trained models" (pre-training), specification of the "training setup", "query pooling", and "joint pre-training". The data analysis stage includes a detailed evaluation of model performance and efficiency, covering aspects such as "model deployment cost", "document representation length", "projections cost", and "projection heads". The workflow culminates in the development of the "improved re-ranking with MVA" model, the study's primary contribution.

*4.5 Trends and Evolution of NLP Research Workflows*

In this section, we investigate the trends and evolution of NLP research workflows to systematically analyze methodological transformations in the field. By statistically evaluating workflow patterns extracted from an extensive corpus of NLP papers, we investigate (1) annual growth trends, (2) distribution across different research stages, and (3) longitudinal changes in the Top 10 most prevalent workflows.

4.5.1 Annual Trends in NLP Research Workflow Counts

---

[6] https://mermaid.js.org

We performed a statistical analysis on the automatically extracted research workflows from NLP papers, correlating them with publication years. This analysis examined both the annual count of unique research workflows per year and the average number of workflows per paper. As shown in **Figure 8**, these two metrics are presented using dual vertical axes due to their differing magnitudes: the left axis indicates the total unique workflows, while the right axis shows the average per paper.

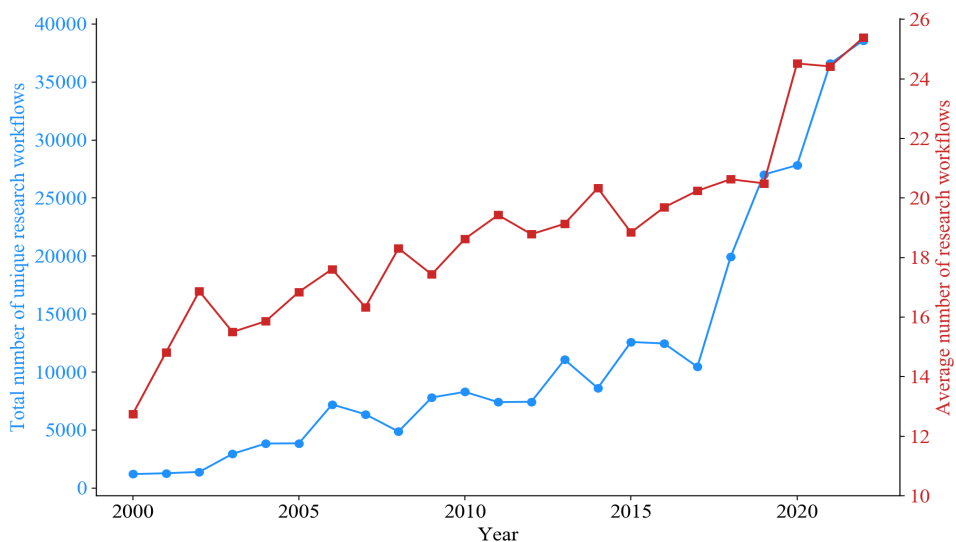

**Figure 8. Growth trends in NLP research workflows (2000–2022)**

The total number of unique NLP research workflows exhibited consistent growth from 2000 (1,206) to 2015 (12,593), followed by a plateau in 2016 (12,468). A marked decline occurred in 2017, attributable to the absence of the NAACL conference proceedings that year, which reduced paper publications and derived research workflows. Post-2018, the field's rapid expansion drove workflow counts to 38,589 by 2022. Regarding workflow frequency per paper, the average increased steadily from 12.8 (2000) to 16.9 (2002), then gradually rose to 20.5 by 2019. A notable surge occurred in 2020 (24.5), culminating at 25.4 in 2022, reflecting intensified methodological diversity in NLP research.

4.5.2 Trends in Workflow Distribution Across Research Stages

NLP research is inherently data-driven, and its typical workflows can be categorized into three main stages: data preparation, data processing, and data analysis. The relative emphasis placed on each stage can be partially inferred from the number of distinct workflow steps identified in that stage. To explore how research efforts are distributed and how these patterns have evolved, we analyzed the average number of workflow phrases per stage across papers published annually, as illustrated in **Figure 9**.

As observed in **Figure 9**, the data processing stage consistently exhibits the highest workflow count across most years, followed by data analysis, while data preparation remains significantly lower. Prior to 2020, data processing and analysis maintained parallel trends, with the former consistently exceeding the latter by 2-3 workflows. However, a distinct shift occurred post-2020: data processing workflows began declining steadily while analysis workflows surpassed processing and continued rising. This reversal suggests a growing NLP research priorities toward interpreting experimental results and investigating the underlying causes of observed phenomena. Conversely, data preparation workflows show a gradual but stable increase, typically ranging between 1-2 workflows annually. This limited variation likely reflects the inherently narrower scope of preparation tasks compared to processing and analysis. Additionally, research papers tend to describe data collection and preprocessing more concisely while dedicating greater detail to subsequent stages.

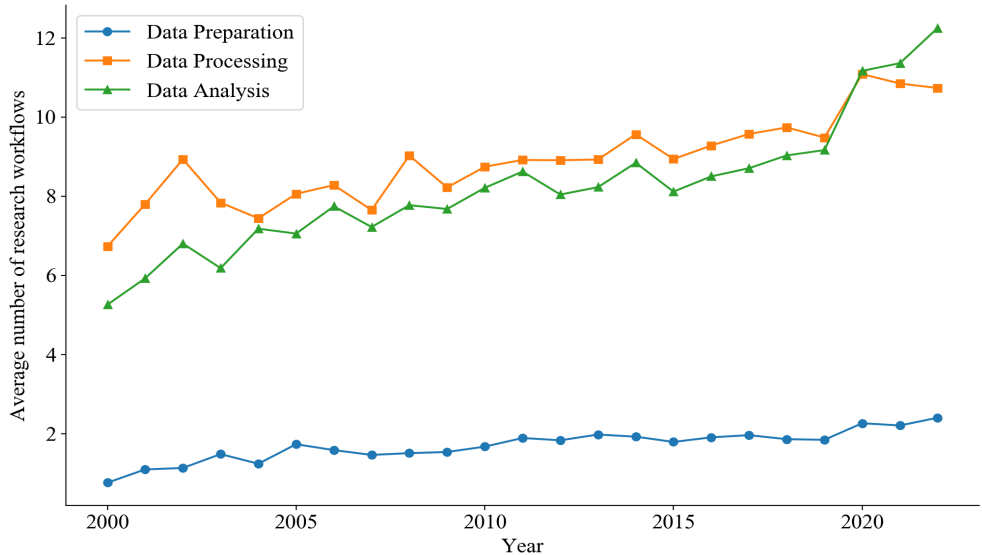

**Figure 9. Temporal trends in average workflow counts across research stages**

4.5.3 Evolution of Top 10 Research Workflows in NLP

To capture the evolving landscape of NLP research, we analyzed the most prevalent research workflows from 2000 onward. For each year, we identified the Top 10 workflows based on their frequency across published papers and calculated their relative proportions within the Top group. **Figure 10** presents the temporal evolution of these workflows, with the horizontal axis indicating the percentage of papers associated with each workflow and the vertical axis showing the publication year. In total, 27 distinct workflows have appeared in the Top 10 over the observation period.

Before 2014, "error analysis" and "case study" consistently ranked as the top two workflows, remaining within the top five thereafter. In 2000, "case study" dominated with 43.42% of papers but declined steadily to 14.10% by 2017, when it was surpassed by "dataset and preprocessing" (15.97%). Over time, the distribution shifted from a "bipolar" structure—where only 2–3 workflows exceeded 10%—to a more diverse "multipolar" pattern, with four or more workflows surpassing 10% annually since 2013, reaching seven in 2022.

In addition to the three aforementioned workflows, "qualitative analysis", "human evaluation", "comparison with baselines", and "data collection" have consistently ranked in the Top 10. The sustained presence of "qualitative analysis" and "human evaluation" highlights the enduring relevance of human-centric evaluation methods, which have gained traction in recent years. The increasing proportion of "comparison with baselines"—exceeding 10% consistently since 2009—reflects the growing emphasis on performance benchmarking. "Data collection", a foundational stage in NLP research, has maintained stable representation (5–9%) throughout the period.

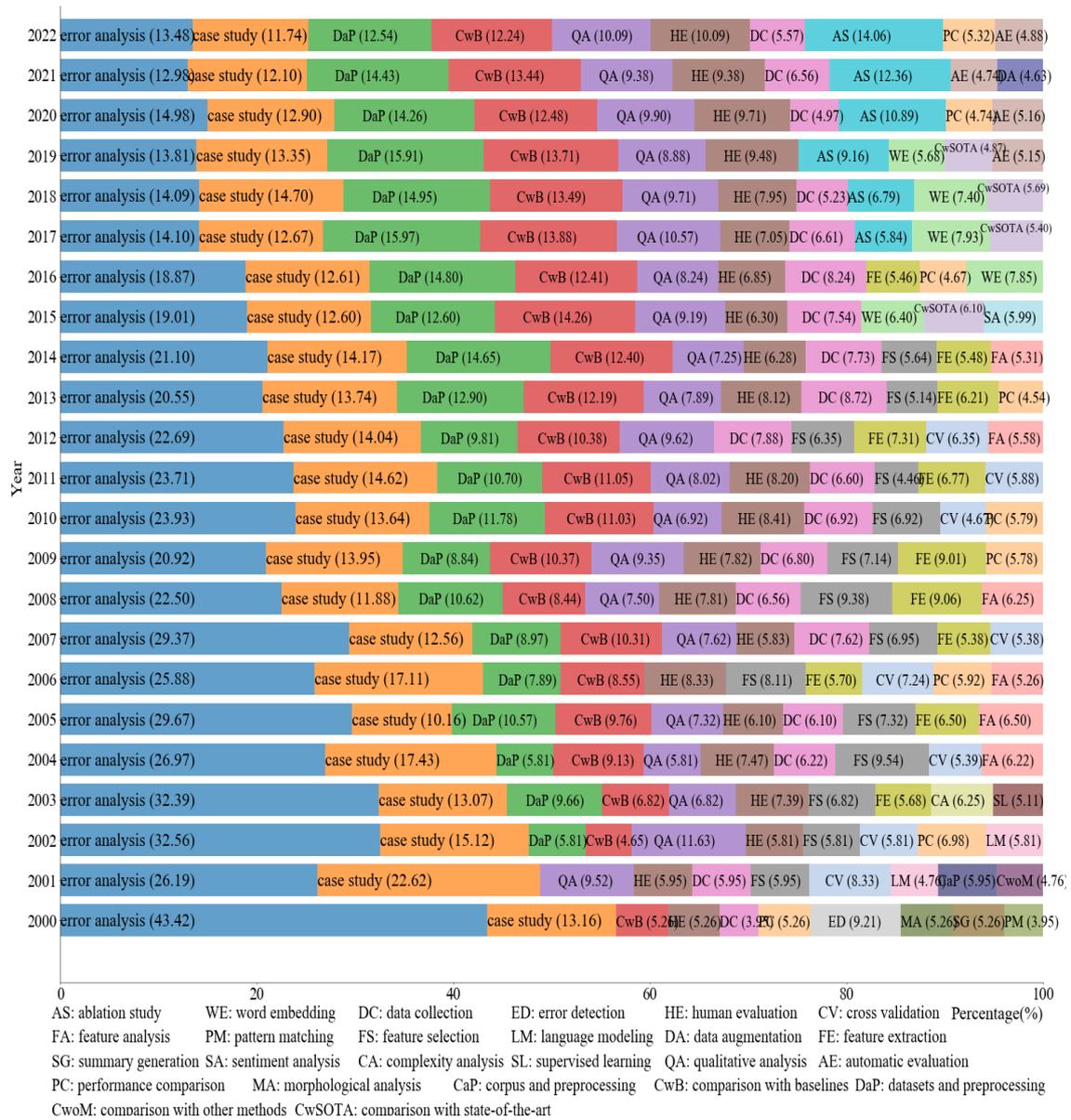

**Figure 10. Temporal evolution of the Top 10 research workflows**

"Ablation study", crucial for interpreting complex deep learning models, first entered the Top 10 in 2017 (5.84%) and rose to become the most prevalent workflow by 2022 (14.06%). In contrast, traditional feature engineering methods like "feature selection" and "feature extraction", once pivotal in machine learning, have disappeared from the Top 10 since 2017. This transition mirrors the broader shift toward end-to-end paradigms facilitated by word embeddings and pre-trained language models, which now dominate contemporary NLP research.

# 5 Discussion

*5.1 Implications*

This study offers significant theoretical implications. Firstly, unlike traditional approaches that examine isolated segments of research workflows, we employ academic full-text mining to automatically generate comprehensive research workflows. This enables visualization of the complete research process, encompassing data preparation, processing, and analysis. Secondly, these research workflows capture researchers' problem-solving experiences and cognitive patterns, representing tacit knowledge that is typically difficult to articulate. In contrast, methods and models extracted through entity recognition represent explicit knowledge. Automatically generated workflows thus facilitate investigation of tacit-explicit knowledge integration and transformation, advancing our understanding of knowledge innovation mechanisms. Finally, our methodology supports automated research workflow generation from large paper corpora within specific domains, offering new insights into field development trends.

The practical implications of this study are threefold. First, automatically generated workflows help researchers quickly grasp papers' core research concepts and serve as innovative navigation tools to locate specific research steps within papers, potentially enhancing literature-assisted reading systems. Secondly, the proposed framework for automated workflow generation can be extended to automatically generate topic-specific research workflow templates from large-scale literature corpora. This capability would offer highly valuable guidance for novice and interdisciplinary researchers, providing structured references for designing their research. Thirdly, our framework establishes the technical foundation for evaluating the completeness and rigor of scientific work from a process-centric perspective. By enabling such assessments, it contributes to developing more holistic and robust academic evaluation systems.

*5.2 Limitations*

While the results of this study are encouraging, several limitations should be acknowledged.

First, the adopted definition of "research workflow" is constrained to a subtask-level procedural sequence of research activities. While this representation captures the general flow, it lacks the fine-grained details essential for full comprehension and reproducibility. Key missing elements include the specific inputs and outputs of each step, the resources employed (e.g., datasets, software), and any conditional constraints or decision points that could alter the workflow's trajectory. This conceptual limitation is compounded by the framework's sequential pipeline, which introduces a susceptibility to cascading errors. Furthermore, the current approach relies predominantly on textual information, neglecting other potentially informative elements within research papers, such as figures, tables, or equations, which could offer complementary insights.

Second, while the general applicability of our framework outside the NLP domain remains to be validated, each individual component has undergone rigorous performance evaluation. The primary limitation lies in the dataset construction process, where we adopted the cost-effective approach commonly employed in NLP literature—using bolded headings to structure methodological procedure descriptions. We designed a streamlined annotation task: annotators only needed to perform a binary classification on whether a given heading could serve as a standalone workflow phrase and if the paragraph content was consistent. This approach significantly reduced manual effort, as it avoided the laborious process of having annotators read full paragraphs and manually summarize a workflow phrase from scratch. However, this methodological convenience creates a dependency on a stylistic feature that may be less prevalent in other academic disciplines. Applying our framework to fields with different writing norms would thus require a more intensive and costly annotation process, where workflow phrases would likely need to be manually authored based on paragraph content. Additionally, the three-stage categorization of data preparation, processing, and analysis is derived from the NLP research paradigm and would require domain-specific adaptation to be applicable elsewhere.

Finally, while Flan-T5 combined with prompt learning yielded the best performance in workflow phrase generation, there remains considerable room for improvement. The achieved ROUGE scores, though demonstrating the feasibility of our approach, are modest. This performance level may reflect the limitations of our pragmatic prompt engineering strategy, which prioritized simple and direct instructions. Although this approach proved both effective and easily reproducible, future work could investigate more advanced techniques. For example, employing more sophisticated prompting methods (e.g., Chain-of-Thought reasoning or automatic prompt optimization), conducting fine-tuning of large language models, or integrating retrieval-augmented generation (RAG) could lead to substantial performance gains. Moreover, the evaluation is constrained by its reliance on ROUGE metrics. These metrics primarily assess n-gram overlap and fail to capture critical dimensions of a high-quality workflow description, such as semantic equivalence, broader grammatical correctness, and overall clarity and conciseness.

## 6 Conclusion and Future Works

This study introduced an innovative framework for automatically generating complete research workflows from full-text academic papers, overcoming the limitations of existing methods that extract only fragmented procedural information. Through developing and validating a multi-stage framework on an extensive corpus of NLP conference papers (2000-2022), we demonstrate a robust and scalable solution for converting unstructured scientific text into structured, visualizable research processes. Our work not only provides a technical solution but also establishes a novel empirical framework for analyzing scientific discipline evolution.

The methodology implements this vision through a three-stage pipeline, with the performance of each stage rigorously validated. For the first step of identifying workflow-descriptive paragraphs, our PU Learning approach combined with SciBERT model achieved an outstanding $F_1$ score of 0.9772, confirming its superior capability in extracting relevant content. Subsequently, for generating concise workflow phrases, a Flan-T5 model enhanced with carefully designed prompt learning yielded optimal results, with, ROUGE-1, ROUGE-2 and ROUGE-L scores of 0.4543, 0.2877 and 0.4427, respectively. Finally, utilizing ChatGPT's few-shot learning capabilities, we accurately classified these phrases into data preparation, processing, and analysis stages (precision = 0.958), enabling structured workflow representations. The seamless integration of these advanced NLP techniques constitutes a major contribution of this work.

Beyond technical validation, applying our framework to over 17,000 NLP papers yielded significant insights into the field's evolution. Our analysis reveals a clear trend toward increasing methodological complexity, with the average workflow steps per paper rising from 12.8 (2000) to 25.4 (2022). We also identified a pivotal shift around 2020, when data analysis workflows surpassed data processing steps, reflecting growing emphasis on model interpretation and result validation. The evolution of Top 10 workflows further illustrates this paradigm shift, showcasing the rise of techniques like "ablation study" and the decline of traditional "feature selection"—directly mirroring the NLP community's transition toward end-to-end deep learning models. These findings demonstrate our approach's unique capability to reveal macro-level scientific domain dynamics. Fundamentally, this research advances research reproducibility, offers an innovative literature navigation system, and establishes a process-oriented foundation for scientific evaluation.

Our future research will focus on the following directions. First, we will improve workflow generation by fine-tuning large language models to produce more detailed and accurate descriptions.

Next, the improved framework will be applied to other academic domains to validate its robustness and facilitate a deeper, comparative analysis of research practices. This will allow us to explore the evolution of research paradigms and the cross-disciplinary propagation of innovative methods. As a long-term objective, we aim to develop automated methods for evaluating research workflow rigor from a process-centric perspective. By training models to assess workflow completeness and adherence to best practices, we seek to create a novel tool that supports peer review, guides researchers, and enhances the systematic evaluation of scientific integrity.

## Acknowledgements

This paper was supported by the National Natural Science Foundation of China (Grant No.72074113, 72374103).

## Author conflict statement

The author(s) declared no potential conflicts of interest with respect to the research, authorship, and/or publication of this article.